\documentclass[times]{TRR}

\usepackage{moreverb,url}
\usepackage{xcolor}
\usepackage[colorlinks,bookmarksopen,bookmarksnumbered,citecolor=red,urlcolor=red]{hyperref}
\usepackage{threeparttable}
\usepackage{multirow}
\newcommand\BibTeX{{\rmfamily B\kern-.05em \textsc{i\kern-.025em b}\kern-.08em
T\kern-.1667em\lower.7ex\hbox{E}\kern-.125emX}}

\def\volumeyear{2022}

\begin{document}

\runninghead{Zhang et al.}

\title{A Multitask Deep Learning Model for Parsing Bridge Elements and Segmenting Defect in Bridge Inspection Images}

\author{Chenyu Zhang\affilnum{1}, Muhammad Monjurul Karim\affilnum{1}, and Ruwen Qin\affilnum{1}}

\affiliation{\affilnum{1}Department of Civil Engineering, Stony Brook University, Stony Brook, NY}

\corrauth{Ruwen Qin, ruwen.qin@stonybrook.edu}

\begin{abstract}
The vast network of bridges in the United States raises a high requirement for maintenance and rehabilitation. The massive cost of manual visual inspection to assess bridge conditions is a burden to some extent. Advanced robots have been leveraged to automate inspection data collection. Automating the segmentations of multiclass elements and surface defects on the elements in the large volume of inspection image data would facilitate an efficient and effective assessment of the bridge condition. Training separate single-task networks for element parsing (i.e., semantic segmentation of multiclass elements) and defect segmentation fails to incorporate the close connection between these two tasks. Both recognizable structural elements and apparent surface defects are present in the inspection images. This paper is motivated to develop a multitask deep learning model that fully utilizes such interdependence between bridge elements and defects to boost the model's task performance and generalization. Furthermore, the study investigated the effectiveness of the proposed model designs for improving task performance, including feature decomposition, cross-talk sharing, and multi-objective loss function. A dataset with pixel-level labels of bridge elements and corrosion was developed for model training and testing. Quantitative and qualitative results from evaluating the developed multitask deep model demonstrate its advantages over the single-task-based model not only in performance (2.59\% higher mIoU on bridge parsing and 1.65\% on corrosion segmentation) but also in computational time and implementation capability.

\end{abstract}

\maketitle

\section{Introduction}
The United States has 619,588 bridges in its inventory, and 36\% need replacement or rehabilitation~\cite{ARTBA}. As the average age of the bridges has increased to 44 years, the cumulative effects of structure loading, environment, and deferred maintenance will deteriorate structural elements progressively~\cite{card}. Bridges in the U.S. are evaluated periodically to determine whether deterioration has occurred, which requires follow-up corrective actions. The conventional manual bridge inspection is still the main form of assessing civil infrastructure conditions~\cite{inspection}. Limitations of the traditional approach motivate research into automating bridge inspection and monitoring using emerging technologies such as robotics and artificial intelligence. It is an inevitable step forward and can be readily adapted to aid and eventually replace manual inspection while offering new advantages and opportunities.

Mobile robots, such as unmanned aerial vehicles (UAVs), are cost-effective tools for gathering inspection video or image data~\cite{spencer2019advances}. They reduce the time, labor, and equipment costs of the onsite inspection process. To make the most of robotic inspection platforms and automation of the inspection process, efficient and dependable methods to analyze inspection data are required. For example, a real-time assessment of the bridge condition based on RGB camera sensors would help prioritize the data collection by other sensors that are in high resolution but more time-consuming and energy-consuming. The Bridge Element Inspection Manual published by the American Association of State Highway and Transportation Officials (AASHTO) \cite{aashto2019manual} states that an overall condition rating of a bridge is based on a comprehensive assessment that relates the defects to specific structural elements. Therefore, it is essential to extract bridge elements and their defects from collected image data.

\begin{figure*}[htbp]
    \centering
    \includegraphics[width=5.5in]{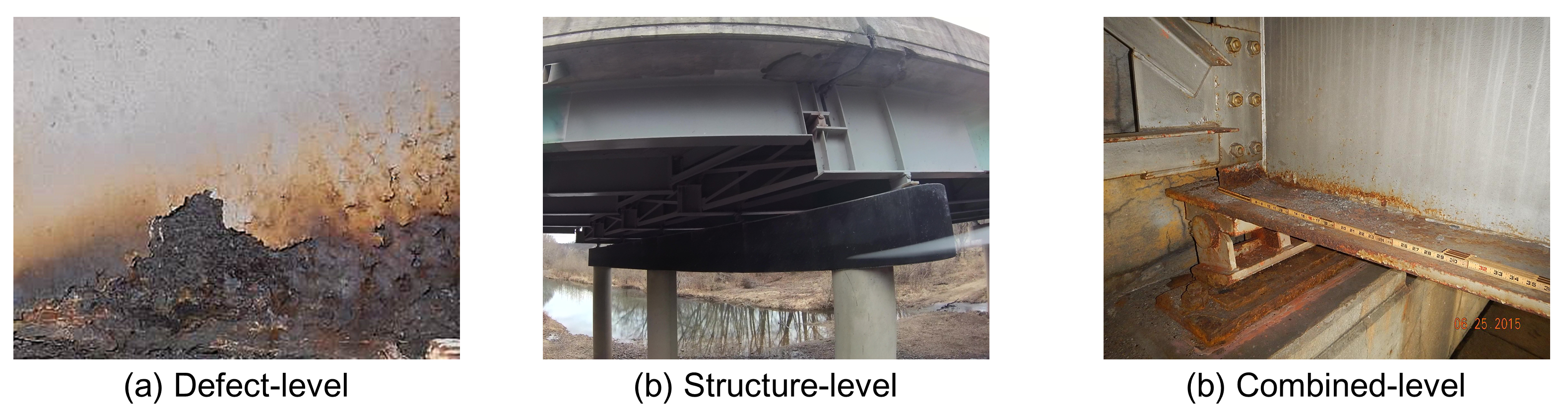}
    \caption{Different Levels of Inspection Images}
    \label{fig:image_level}
\end{figure*}

Detecting and segmenting defects on civil structures is an area of research interest in structural health monitoring. The rapid development of graphics processing units (GPUs) and deep learning algorithms have significantly advanced computer vision-based defect analysis. Researchers have achieved impressive results in identifying various damage or defect types, such as concrete cracks, steel cracks, asphalt cracks, spalling, steel corrosion, exposed rebar, and many others~\cite{bianchi2022visual}. However, many models are trained on images that primarily contain pixel-level details of surface defects and can be called defect-level images, as shown in Figure~\ref{fig:image_level}(a). Those images are relatively easy to analyze because defects are in a closer look, and the background is greatly homogeneous. Defect-level images must be either taken from a position closer to defects or pre-processed (i.e., cropped and zoomed in), which raises higher requirements for the UAV's operator or increases the workload of image analysis.

Compared to the large volume of studies on defect detection and segmentation, research on extracting multiclass structural elements from inspection images or videos is limited and remains challenging \cite{narazaki2020vision, bianchi2021coco, karim2022semi, zhang2022adeep}. Prior work used structure-level images where multiple elements of a civil structure are salient objects, as shown in Figure~\ref{fig:image_level}(b). Many elements and components are distinguishable in these images, but the detailed information on the element surface is missing.

Nevertheless, cameras also capture bridge images from a certain distance where images have recognizable structural elements and apparent surface defects on the elements, as shown in Figure \ref{fig:image_level}(c). Combined-level inspection images like such provide valuable information for a quick, automatic assessment of the bridge condition. In those images, bridge elements of the same type have widely different appearances due to the imperfect and changing view of the mobile cameras. Due to the typically irregular shapes of bridge elements and defects, semantic segmentation is more suitable than other identification approaches for parsing inspection images. To our best knowledge, this kind of inspection image and the association between bridge elements and their defects have not been fully utilized and studied yet.

A naive way to analyze bridge elements and defects in combined-level images is to train separate networks for the two tasks. However, this approach ignores the relationship between these two tasks. It is likely to obtain low generalizability and learn similar features resulting in inefficient use of information-sharing, computation resources, and network capability. Attempts to improve the performance of element and defect segmentation have mainly focused on refining network architectures (deeper and broader) and increasing training data. Instead, an efficient approach would be multitask learning (MTL), which learns multiple tasks simultaneously by optimizing more than one loss function in a single model. An MTL model learns generalized data representations, which are helpful in multiple contexts. Besides improved learning efficiency and prediction accuracy, MTL can reduce the risk of overfitting.

The task of structural element parsing involves extracting structural elements of bridges and classifying them. The defect segmentation task is to assign a defect label for each pixel. Structural element parsing and defect segmentation are two correlated tasks. For example, corrosion can be developed on elements made with metals but not concrete, masonry, and composite materials. This correlation has not been fully utilized and studied yet. Given the hypothesis that related tasks may benefit from MTL, this paper proposes a multitask deep learning model to parse multiclass bridge elements and segment defects simultaneously. Aligning the two tasks' outputs from the multitasking model is convenient and allows easily locating of both defective elements in images and the defects on those elements.

The rest of the paper is organized as follows. The next section summarizes the related work. Then, the following section introduces the proposed MTL network and evaluation metrics. After that, the dataset and implementation details are delineated, followed by a quantitative and qualitative analysis of the results. Finally, the conclusion and future work are summarized.

\section{Literature Review}

The literature related to this study includes computer vision-based defect segmentation, structural element segmentation, and MTL. 

\subsection{Defect Segmentation}

Many prior studies concentrated on locating concrete cracks using conventional image processing methods. Restricted by limited computational resources, these methods can only identify defects with specific patterns without utilizing the contextual information in inspection images. The performance of defect detection improved significantly with the popularity of convolutional neural networks (CNNs) \cite{zhang2016road} and the drop in computational costs. The high-resolution network (HRNet) \cite{wang2020deep} is a general-purpose CNN that can maintain high-resolution representations by connecting high-to-low-resolution convolutions in parallel. It has been successfully applied to surface defect detection and segmentation \cite{yudin2020roof, jia2021surface, akhyar2021beneficial}.

Both semantic segmentation and instance segmentation can be used to locate defects in inspection images. Semantic segmentation classifies every pixel in an image without differentiating objects of the same class in an image. Rubio et al. \cite{rubio2019multi} and Dung et al. \cite{ dung2019autonomous} performed semantic segmentation of bridge deck damages (delamination and rebar exposure) and concrete cracks, respectively, using a fully convolutional network. Based on the U-Net \cite{ronneberger2015u} and attention mechanism, Pan et al. \cite{pan2020automatic} developed a PipeUNet to segment sewer defects (crack, infiltration, joint offset, and intruding lateral). Wang and Cheng \cite{wang2020unified} proposed a DilaSeg-CRF, which is based on PSPNet \cite{zhao2017pyramid}, for segmenting sewer pipe defects (crack, deposit, and root). Shi et al. \cite{shi2021improvement} segmented steel corrosion and rubber bearing crack using a U-Net. Instance segmentation, however, is a task combining semantic segmentation and object detection, which not only classifies each pixel but differentiates different objects of the same class. Mask R-CNN \cite{he2017mask} is a deep learning model of instance segmentation, and it has been successfully utilized for segmenting concrete bridge cracks \cite{ayele2020automatic}, façade defects \cite{li2022multi}, and shield tunnel lining cracks \cite{huang2022deep}.

\subsection{Structural Element Segmentation}

Element segmentation is an indispensable part of an automated bridge inspection, yet it has not progressed to a satisfying level. Extracting bridge elements from inspection images is still not completely solved due to its higher complexity than the defect segmentation task and the lack of big inspection datasets.

Yeum et al. \citep{yeum2019automated} developed a CNN-based method to extract the regions of interest in inspection images for a visual evaluation. Narazaki et al. \citep{narazaki2020vision} trained a scene classifier to segment five bridge components. Bianchi et al. \citep{bianchi2021coco} applied the SSD v2 model \cite{liu2016ssd} to bridge details detection and investigated the use of image augmentation to boost the performance of the trained model. Karim et al. \citep{karim2022semi} transferred a Mask R-CNN, pre-trained on a large public dataset, to segment multiclass bridge elements. They also developed a semi-supervised self-training method to engage the inspector to refine the network iteratively. Through a comparison with the state-of-the-art semantic segmentation networks, Zhang et al. \cite{zhang2022adeep} demonstrated that HRNetV2-W32 is particularly suitable for extracting deep features to segment multiclass structural elements from bridge inspection images.

\subsection{Multitask Learning}

Multitask learning (MTL) has been successfully used in many computer vision tasks, such as depth estimation and surface normal prediction \cite{eigen2015predicting}, segmentation of building footprints \cite{bischke2019multi}, pedestrian detection \cite{tian2015pedestrian}, and others. The most commonly used way to achieve MTL is to share the feature extractor and branch downstream tasks for task-wise predictions \cite{ruder2017overview}. The shared feature extractor learns common representations for all tasks, dramatically lowering the overfitting risk and improving generalization \cite{zhang2021survey}. Hoskere et al. \citep{hoskere2020madnet} proposed MaDnet, a deep neural network composed of a shared feature extractor and multiple semantic segmentation paths to identify material and damage types. This framework suggests that contextual information exists and can be helpful to the segmentation tasks. For the bridge element and defect segmentation tasks, one can be an auxiliary task to help another by introducing more annotations and perspectives.

\subsection{Research Gaps}

Nonetheless, current studies on computer vision-based civil infrastructure inspection lack a connection between the defect and structural element analysis. It is well-known to the civil engineering community that structural elements and defects have a spatial correlation. Integrating this prior knowledge into the design of machine learning models for bridge condition assessment would be a valuable step.

\section{Methodology}
\begin{figure*}[htbp]
    \centering
    \includegraphics[width=\textwidth]{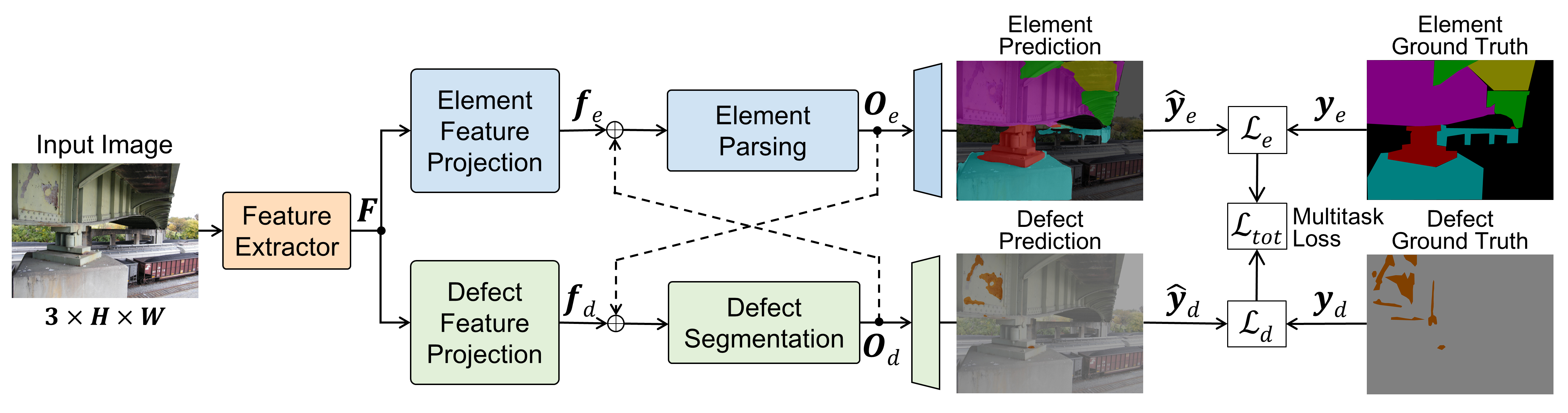}
    \caption{Architecture of the Multitask Deep Learning Model}
    \label{fig:MTL_architecture}
\end{figure*}

Inspired by the correlation between structural elements and defects, this paper proposes a MTL model to parse structural elements and segment defects in parallel. Figure \ref{fig:MTL_architecture} illustrates its architecture. The input is an RGB (red, green, and blue) three-channel image with height $H$ and width $W$. First, the shared feature extractor turns the input image into a common feature embedding, $\pmb{F}(\in\mathbb{R}^{c\times h\times w})$, which consists of $c$ feature maps with height $h$ and width $w$. Then, $\pmb{F}$ flows into two branches that are respectively dedicated to the two tasks: element parsing and defect segmentation. 

In the element parsing branch, a projection network extracts the element-specific feature embedding, $\pmb{f}_e(\in\mathbb{R}^{c \times h \times w})$, from the common feature embedding $\pmb{F}$. $\pmb{f}_e$ flows through the element parsing network to become a downsampled feature embedding, $\pmb{O}_e (\in\mathbb{R}^{c_e \times h \times w})$, with the number of channels equal to the number of element classes, $c_e$. $\pmb{O}_e$ is upsampled to become the element prediction, $\hat{\pmb{y}}_e (\in\mathbb{R}^{c_e \times H \times W})$. The branch of the defect segmentation is similarly defined.

The output from the defect segmentation network, $\pmb{O}_d(\in\mathbb{R}^{c_d\times h \times w})$, can be concatenated with the element-specific feature embedding $\pmb{f}_e$ to provide the defect awareness to the element parsing task. Similarly, $\pmb{O}_e$ can provide the element awareness to the defect segmentation task. This is termed cross-talk sharing. 
The dashed lines in Figure \ref{fig:MTL_architecture} represent this design, which will be evaluated in the result section. This section further discusses the details of model components and the development method below.

\subsection{Feature Extraction}

Semantic segmentation is a highly position-sensitive task that desires spatially-fine representation. Therefore, the proposed model uses HRNet-W32 as its feature extractor that turns an input image into its feature embedding. This network can maintain high-resolution representations by connecting high-to-low resolution convolutions in parallel. In this study, the dimensions of input images are $3\times 480 \times 480$ (i.e., $H=W=480$). The common feature embedding $\pmb{F}$ of an input image contains 480 feature maps in size 120$\times 120$ (i.e., $c=480$, and $h=w=120$).

\subsection{Feature Decomposition}

The common feature embedding of an input image, $\pmb{F}$, contains both element-related and defect-related information. The element-specific features, $\pmb{f}_{e}$, and the defect-specific features, $\pmb{f}_{d}$, intertwined in $\pmb{F}$ are decomposed by two projection modules:

\begin{equation}
    \begin{aligned}
    &\pmb{f}_{e} = P(\pmb{F}; \pmb{\Phi}_{e}, \pmb{\beta}_{e})\\
    &\pmb{f}_{d} = P(\pmb{F}; \pmb{\Phi}_{d}, \pmb{\beta}_{d})\\
    \end{aligned}
\end{equation}
where $P(\pmb{F};\pmb{\Phi},\pmb{\beta})$ represents a projection network that maps the input feature map $\pmb{F}$ onto a new space in the same dimension using the projection matrix $\pmb{\Phi}(\in\mathbb{R}^{c\times c})$ and the bias vector $\pmb{\beta}(\in\mathbb{R}^{c\times 1})$. However, the projection network can have different designs. For example, When both $\pmb{\Phi}$ and $\pmb{\beta}$ are scalars, the projection is simply about scaling the input embedding linearly and then shifting it to become the task-specific feature embedding. When the projection parameters are vectors in the same dimension as the input embedding's feature map numbers ($c$), the projection is about adding the channel-wise attention to $\pmb{F}$. The result section will evaluate the performance of the different projection designs.

\subsection{Segmentation Networks}
\if
In each branch, one convolutional network is used to downsample the feature embedding $\pmb{f}$ to become $\pmb{O}$. The information flow for both the branches can be represented as:

\begin{equation}
    \begin{aligned}
    &\pmb{O}_e = \text{Conv}(\text{Conv}(\text{Cat}(\pmb{f}_e,\pmb{O}_d); \pmb{\theta}_{e1});\pmb{\theta}_{e2})\\
    &\pmb{O}_d = \text{Conv}(\text{Conv}(\text{Cat}(\pmb{f}_d,\pmb{O}_e); \pmb{\theta}_{d1});\pmb{\theta}_{d2})\\
    \end{aligned}
\end{equation}
where Cat($\cdot$,$\cdot$) means the concatenation operation, Conv($\cdot$;$\pmb{\theta}$) means convolution operation with learnable parameters $\pmb{\theta}$. In this study, $\pmb{O}_e\in \mathbb{R}^{7\times 120 \times 120}$ (i.e., $c_e=7$) since the element parsing task has 7 classes including the background class. $\pmb{O}_d\in \mathbb{R}^{2\times 120 \times 120}$ (i.e., $c_d=2$) because the defect segmentation has only two classes (corrosion vs. no corrosion).

The downsampled feature embeddings, $\pmb{O}_e$ and $\pmb{O}_d$, are projected into two separate score maps to get the element and defect predictions, $\widehat{\pmb{y}}_e$ and $\widehat{\pmb{y}}_d$, respectively:

\begin{equation}
    \begin{aligned}
    & \widehat{\pmb{y}}_e = S(\pmb{O}_e; \pmb{\varphi}_e) \\
    & \widehat{\pmb{y}}_d = S(\pmb{O}_d; \pmb{\varphi}_d) \\
    \end{aligned}
\end{equation}
where $S(\cdot ; \pmb{\varphi})$ represents the mapping function with learnable parameters $\pmb{\varphi}$ that upsamples the output maps to become the same size as the input images. 
This mapping function upsamples the output score maps to become the same size of the input images' label maps. That is, $\widehat{\pmb{y}}_e\in\mathbb{R}^{7 \times 480 \times 480}$ and $\widehat{\pmb{y}}_d\in\mathbb{R}^{2 \times 480 \times 480}$. 
\fi

For both branches, a convolutional network downsamples the feature embedding $\pmb{f}$ to become the class-specific feature maps $\pmb{O}$. The information flow for the two classification branches can be represented as:

\begin{equation}
    \begin{aligned}
    &\pmb{O}_e = \text{Conv}(\text{Conv}(\text{Cat}(\pmb{f}_e,\pmb{O}_d); \pmb{\theta}_{e1});\pmb{\theta}_{e2})\\
    &\pmb{O}_d = \text{Conv}(\text{Conv}(\text{Cat}(\pmb{f}_d,\pmb{O}_e); \pmb{\theta}_{d1});\pmb{\theta}_{d2})\\
    \end{aligned}
    \label{eq:classification}
\end{equation}
where Cat($\cdot$,$\cdot$) means the concatenation operation that links tensors together along a given dimension, Conv($\cdot$; $\pmb{\theta}$) means convolution operation with learnable parameters $\pmb{\theta}$. In this study, $\pmb{O}_{e}\in \mathbb{R}^{7\times 120 \times 120}$ (i.e., $c_e=7$) since the element parsing task has 7 classes including the background class. $\pmb{O}_{d}\in \mathbb{R}^{2\times 120 \times 120}$ (i.e., $c_d=2$) because the defect segmentation has only two classes (corrosion vs. no corrosion). $\pmb{O}_d$ and $\pmb{O}_e$ in equation (\ref{eq:classification}) are the feature maps saved at the last time step of model training, as the dashed lines in Figure \ref{fig:MTL_architecture} indicate. In the first step, $\pmb{O}_d$ and $\pmb{O}_e$ use identify matrices as their initialization.


The downsampled feature maps, $\pmb{O}_{e}$ and $\pmb{O}_{d}$, are projected into two separate score maps to get the predictions of element and defect classes, $\widehat{\pmb{y}}_{e}$ and $\widehat{\pmb{y}}_{d}$, respectively:

\begin{equation}
    \begin{aligned}
    & \widehat{\pmb{y}}_e = S(\pmb{O}_e; \pmb{\varphi}_e) \\
    & \widehat{\pmb{y}}_d = S(\pmb{O}_d; \pmb{\varphi}_d) \\
    \end{aligned}
\end{equation}
where $S(\cdot ; \pmb{\varphi})$ represents the mapping function with learnable parameters $\pmb{\varphi}$ that upsamples the feature maps to become the same size as input images. 
That is, $\widehat{\pmb{y}}_{e}\in\mathbb{R}^{7 \times 480 \times 480}$ and $\widehat{\pmb{y}}_{d}\in\mathbb{R}^{2 \times 480 \times 480}$.

\subsection{The Loss Function}

The cross-entropy loss for the element parsing task, $\mathcal{L}_e$, and that for the defect segmentation task, $\mathcal{L}_d$, are calculated:

  \begin{equation}
  \begin{aligned}
  &\mathcal{L}_e = - \sum_{k}<\pmb{y}_e(k), \log\widehat{\pmb{y}}_{e}(k)>\\
  &\mathcal{L}_d = - \sum_{k}<\pmb{y}_d(k), \log\widehat{\pmb{y}}_{d}(k)>
  \end{aligned}
  \label{eq:losses}
  \end{equation}
to measure the errors of prediction scores compared to the true classes. $k$ in equation (\ref{eq:losses}) is the index of images in the training dataset, $\pmb{y}(k)$ and $\widehat{\pmb{y}}(k)$ are the ground-truth and prediction of the $k$th training image, and $<,>$ is the Frobenius inner product. 

The respective loss functions of the two tasks are combined to become the multitask loss function, $\mathcal{L}_{tot}$. The easiest way is to simply use equal weights for combining the tasks, $\mathcal{L}_e+\mathcal{L}_d$, called the additive loss function. However, the performance of MTL strongly depends on the relative weights between each task's loss~\cite{gong2019comparison}. 
Manually tuning these weights is difficult and costly, making MTL prohibitive in practice. Therefore, developing a more practical method to determine the optimal weights would be ideal. One approach considered in this study is to use the homoscedastic uncertainty of each task \cite{kendall2018multi} to weight its loss:

  \begin{equation}
  \mathcal{L}_{tot} = \frac{1}{2\sigma_e^2 } \mathcal{L}_e + \frac{1}{2\sigma_d^2 } \mathcal{L}_d + \log\sigma _e\sigma _d
  \end{equation}
where $\sigma_e$ and $\sigma_e$ are noise parameters for the two respective tasks. They are learnable weights for the minimization objective. Both the additive and homoscedastic uncertainty loss functions are examined in the result section.

\subsection{Evaluation Metrics}

To evaluate the performance of the two segmentation tasks, four metrics are calculated at the class level for each task:

\begin{equation}
     \text{Precision}=\frac{\text{No. correctly predicted pixels}}{\text{No. pixels predicted to be the class}}
\end{equation}

\begin{equation}
    \text{Recall}=\frac{\text{No. correctly predicted pixels}}{\text{No. pixels of the class}}
\end{equation}

\begin{equation}
    \text{F1}=\frac{2\times\text{Precision}\times \text{Recall}}{\text{Precision}+\text{Recall}}
\end{equation}

\begin{equation}
    \text{IoU} = \frac{\text{No. pixels of (ground-truth $\cap$ prediction masks)}}{\text{No. pixels of (ground-truth $\cup$ prediction masks)}}
\end{equation}
where IoU stands for Intersection over Union. 

Then, the average of class-level values is further calculated for each of the four metrics to obtain the task-level performance metrics, including mean Precision (mPrecision), mean Recall (mRecall), mean F1-score (mF1), and mean IoU (mIoU).

\section{Dataset and Implementation}

The proposed MTL model requires labeled data for training and testing. This section illustrates the dataset developed for this study and details of model implementation.

\subsection{The Dataset}
As one of the most common types of bridges in the U.S., steel bridges are susceptible to atmospheric corrosion. Deck runoff and traffic-generated aerosols that contain deicing salts can accelerate corrosion. General corrosion may result in loss-of-section of bridge components, greater dead, live stresses, and stress concentrators, which might lead to cracking issues in the future. Today, general corrosion on steel bridges is perhaps the most prevalent issue.

The study covered 145 images, a portion of the Corrosion Condition State Semantic Segmentation Dataset \cite{Bianchi2021}. The inspection images were collected from the Virginia Department of Transportation Bridge Inspection Reports and semantically annotated following the corrosion condition state guidelines stated in the Bridge Inspector's Reference Manual~\cite{aashto2019manual}. 

This project used the image annotation tool LabelMe \cite{russell2008labelme} to provide the pixel-level annotation of six common classes of structural elements for steel girder bridges in these inspection images: Bearing (Brg), Bracing (Brc), Deck (Dck), Floor beam (Flb), Girder (Grd), and Substructure (Sbt). In total, 822 instances are annotated in the 145 images. Also, the study converted the annotation of the original four corrosion class categories: Good, Fair, Poor, and Severe, into the annotation of binary classes: Corrosion (Cor) and No Corrosion (No), for the MTL model. The study reserves 130 images for training models and 15 images for testing. Table \ref{tab:data} further summarizes the distribution of element instances and the pixel proportion of defect areas (\% defect area) in the training and testing subset. The structural element proportions range from 10\% to 29\% and are all above 10\%. The defect area proportions are about 2:8. The distributions reflect the actual bridge inspection situation, indicating that the dataset is eligible for developing the proposed deep learning framework.

\begin{table*}[!ht]
	\caption{Distribution of the Structural Elements and Defect}\label{tab:data}
	\begin{center}
        \begin{tabular}{l| r| r r r r r r| r r}
        \hline
        \multirow{2}{*}{}&\multicolumn{1}{c|}{No.}&\multicolumn{6}{c|}{No. structural elements}&\multicolumn{2}{c}{\% defect area}\\
        \cline{3-10}
          & images  & Brg  & Brc  &  Dck  & Flb & Grd & Sbt &Cor & No\\
        \hline
        Training & 130  & 169  & 76  &  104  & 84 & 217 & 99 & 17.3 & 82.7 \\
        \;\!Testing  & 15  & 19  & 6  &  6  & 10 & 20 & 12 & 13.9 & 86.1 \\
        \rule{-2pt}{10pt}
        Total  & 145 & 188  & 82  & 110  & 94 & 237 & 111 & 16.9 & 83.1\\
        \hline
        \multicolumn{1}{l}{\,Proportion (\%)}& &23 & 10  & 13  & 11  & 29 & 14 & NA &NA\\
        \hline
		\end{tabular}
	\end{center}
\end{table*}

\subsection{Model Training}

Training a deep learning model from scratch requires big data, which is unrealistic for studies with small data. Transfer learning, which adapts a model pre-trained on big data to a related task using a small dataset, can effectively address this challenge. In transfer learning, the small data collected for the target task are primarily used to retrain or refine task-specific parameters. The backbone HRNet-W32 in this study was initialized by adopting the weights pre-trained on the PASCAL Visual Object Classes (VOC) 2012 dataset~\cite{pascal-voc-2012} and Semantic Boundaries Dataset (SBD)~\cite{BharathICCV2011}. Then, transfer learning adapted the backbone to the two semantic segmentation tasks. Training and testing were performed using one Nvidia Tesla V100 GPU with 32 GB of memory. For consistency, all images were resized from the original size to $480\times480$ (i.e., $H=W=480$). This study also employed data augmentation techniques to increase the diversity and representation of the training dataset, thus improving the network's generalization ability. Specifically, the data augmentation methods include random scale transformation, image scaling, random zooming, random rotations between $\pm10^{\circ}$ and, random horizontal flipping, random image intensity noise using $5\times5$ Gaussian kernel, and HSV augment that randomly adjusts image hue (H), saturation (S), and value (V). The optimizer Adam was used for training models. The schedule applies an exponential decay function to the optimization step, given an initial learning rate of 5e-4 and the minimum rate of 5e-6. The batch size used in this training process is eight.

\section{Experiments and Results}

This section compares the proposed MT model with the individual single-task networks in a series of experiments. Results from the conducted experiments are discussed.

\subsection{Design of Experiments}

To illustrate the merits of the MTL model, three single-task HRNetV2-W32 models were trained, where each is composed of the backbone HRNet-W32 and a segmentation head. The first two single-task networks were trained for bridge element parsing and defect segmentation, respectively. The third network (merged-task) combines element and defect labels to define bridge element classes as elements with defects and without defects. To fully examine and boost the potential and performance of MTL, this study developed twelve different versions of the MTL model as shown in Table \ref{tab:exp}. Firstly, three kinds of projection method were examined: scalar, vector, and matrix. Then, the effect of cross-talk sharing (i.e., the dashed lines in Figure \ref{fig:MTL_architecture}) between the MTL model's two branches was explored. Furthermore, two different loss functions, the additive loss and the homoscedastic uncertainty loss, were compared.

\begin{table*}[!ht]
	\caption{Different Versions of the Multitask Deep Learning Model}\label{tab:exp}
	\begin{center}
		\begin{tabular}{l| l l l}
		\hline
			Model & Projection & Cross-talk & Loss function \\\hline
			MTL-A   & Scalar & No & Additive \\
			MTL-B   & Scalar & No & Homoscedastic uncertainty \\
			MTL-C   & Scalar & Yes & Additive\\
			MTL-D   & Scalar & Yes & Homoscedastic uncertainty\\\hline
			MTL-E & Vector & No & Additive\\
			MTL-F & Vector & No & Homoscedastic uncertainty \\
			MTL-G & Vector & Yes & Additive\\
			MTL-H  & Vector& Yes & Homoscedastic uncertainty\\\hline
			MTL-I   & Matrix & No & Additive \\
			MTL-J   & Matrix & No & Homoscedastic uncertainty \\
			MTL-K   & Matrix & Yes & Additive\\
			MTL-L   & Matrix & Yes & Homoscedastic uncertainty\\\hline
		\end{tabular}
	\end{center}
\end{table*}

\subsection{Performance Comparison at the Task Level}

The twelve versions of the MTL model (A$\sim$L) are compared with the three single-task networks. Table \ref{tab:perfor} summarizes the segmentation performance measurements in this comparative study.
The two independent single-task networks achieved fine results on their respective task with limited training data. The merged-class single-task network achieved comparable results to the single-task network on the element parsing task. However, its performance on the corrosion segmentation task dropped by 27.12\% on mIoU, 12.92\% on mPrecision, 28.83\% on mRecall, and 21.39\% on mF1. The elements with and without corrosion both lose their shape and position features. Therefore, it becomes difficult to classify the corrosion for the model, resulting in decreased corrosion segmentation performance. Overall, the merged-class single-task network is unsuitable for these two tasks, although it can simultaneously complete two tasks using a single-task architecture.

The twelve versions of the MTL model outperformed the single-task and the merged-class single-task networks for all the performance metrics of the two tasks except for the mRecall of corrosion segmentation task. The improvement in mIoU of the bridge parsing task due to the introduction of MTL ranges from 0.78\% to 2.59\%, and 0.55\% to 2.46\% for the corrosion segmentation task with the same training data and hyperparameters. The MTL-I model is the best in terms of mIoU, mRecall, and mF1 on the bridge parsing task. MTL-D is the best in terms of mIoU and mF1 on the corrosion segmentation task. These two models will be further discussed and compared in the following.

For the projection method, as the complexity increases from a scalar to a matrix, there is no clear improvement. The potential reason could be that more complexity means more difficulty in training and optimization. Then, for the cross-talk sharing between the two tasks, the models were improved, particularly on the corrosion segmentation task. The highest increase in mIoU was 1.21\% between MTL-C and MTL-A for the corrosion task. Last but not least, the modification of the loss function matters in MTL. The MTL models using the scalar projection benefit from the homoscedastic uncertainty weight, as well as the networks using the vector projection without cross-talk. The improvements were not remarkable for the networks using the vector projection with cross-talk and the matrix projection as they have too high complexity to optimize with limited training data. However, the performance of complex models could be further improved by increasing the size of the training dataset.

\begin{table*}[htbp]
	\caption{Performance Comparison Between Single- and Multi-task Networks}\label{tab:perfor}
	\begin{center}
	\begin{threeparttable}
        \begin{tabular}{l|r r r r| r r r r}
        \hline
        \multirow{2}{*}{Network}&\multicolumn{4}{c|}{Bridge Element Parsing Task}&\multicolumn{4}{c}{Corrosion Segmentation Task}\\
        \cline{2-9}
          & mIoU  & mPrecision  & mRecall  &  mF1  & mIoU  & mPrecision  & mRecall  &  mF1\\\hline
        Single-task & 82.89 & 90.75 & 90.41 & 90.58 & 80.85 & 87.20 & 90.76 & 88.94\\
        Merged-task & 82.35&90.17& 90.39&90.28&53.73 &74.28&61.93& 67.55 \\\hline
        MTL-A & 85.12&\textbf{92.34}&91.42 & 91.88 & 81.47 &88.78 & 89.74 & 89.26 \\
        MTL-B & 84.38 & 91.74 & 91.14  & 91.44 & 83.15 & \textbf{90.58} & 90.14 & 90.36\\
        MTL-C & 83.67 & 91.87 & 90.37&91.11&82.61&89.47&90.58&90.02\\
		\textbf{MTL-D} & 84.41 & 92.01 & 91.04& 91.52&\textbf{83.31}&90.09&90.85&\textbf{90.47}\\
		MTL-E & 84.39 & 91.36 & 91.74&91.55&82.78&89.39&\textbf{90.90}&90.14\\
		MTL-F & 84.66&91.63& 91.60&91.61&82.93& 90.01&90.43&90.22\\
		MTL-G & 85.00&91.51&92.13&91.82&82.26&89.04&90.57&89.80\\
		MTL-H &84.82&91.37&92.09&91.73&81.91&88.66&90.51&89.58\\
		\textbf{MTL-I} &\textbf{85.48}&92.00&\textbf{92.33}&\textbf{92.16}&82.50&89.74&90.13&89.93 \\
		MTL-J &83.74&91.63&90.56&91.09& 81.88&88.90&90.20&89.55\\
		MTL-K & 84.63 & 92.12& 91.11& 91.61&82.79&89.94&90.31&90.12\\
		MTL-L & 83.79 & 91.25&90.90&91.07&82.31&89.49&90.13&89.81\\\hline
		\end{tabular}
		\begin{tablenotes}
        \footnotesize
        \item \textit{Note}: The highest value for each index among all networks is shown in bold text.
        \end{tablenotes}
    \end{threeparttable}
	\end{center}
\end{table*}

\subsection{Performance Comparison at the Class Level}

To illustrate the performance improvement at the class level by the introduction of MTL, the two MTL models, MTL-D and MTL-I, performed the best in terms of mIoU on the bridge element parsing task and defect segmentation task, were selected to have a close look as shown in Table \ref{tab:class}. The single-task networks are taken as the benchmark for comparison.

For the bridge element parsing task, the MTL-I model has a noticeable performance improvement compared to the single-task network. It significantly outperforms the single-task network on IoU, particularly for bracing and deck with more than 5\% improvement (6.42\% for bracing and 5.02\% for deck). For substructure and background, the increases in IoU are more than 2\%, which is also inspiring (2.64\% for substructure and 2.40\% for background). For the IoU of segmenting bearing, floor beam, and girder classes, MTL-I still achieves better results, although the improvements are limited to 1\% (0.27\% for bearing, 0.84\% for floor beam, and 0.58\% for girder). In precision, the MTL-I network broadly achieves better results on each class except for bracing, ranging from 0.45\% on the girder class to 3.61\% on the floor beam class. In recall, the improvement by the MTL-I network can be seen in each class apart from bearing and floor beam classes, and especially for bracing, the progress reached surprisingly 10.60\%. The introduction of defect features provides more information to the model and reinforces its segmentation capability on the edges of elements. Therefore, the defect features improve the MTL models' performance on the element parsing task.

For the defect segmentation task, the MTL-D network also shows a dominant performance to the single-task network. In IoU, the improvements of the MTL-D network are 3.78\% for the corrosion class and 1.13\% for the no corrosion class, respectively. In precision, the MTL-D model outperforms with a 5.92\% improvement for the corrosion class and a comparable performance of 97.50\% for the no corrosion class. In recall, it is observed that an increase (1.33\%) and a decrease (1.18\%) on no corrosion and corrosion, respectively. This is because corrosion typically occurs on specific bridge elements that are easily contacted by deck runoff and traffic-generated aerosols. The element features offer hints for the corrosion segmentation task.

All in all, the MTL-I model has a much better performance on the bridge element parsing task and comparable performance on the corrosion segmentation task compared with the MTL-D network. It is recommended to use the MTL-I network in the application and implementation.

\begin{table*}[!ht]
	\caption{Performance comparison by class between single-task and well-performed MTL networks}\label{tab:class}
	\begin{center}
        \begin{threeparttable}
        \begin{tabular}{l|r r r|r r r|r r r}
        \hline
        \multirow{2}{*}{Class}&\multicolumn{3}{c|}{IoU}&\multicolumn{3}{c|}{Precision}&\multicolumn{3}{c}{Recall}\\
        \cline{2-10}
        & Single  & MTL-D  & MTL-I  &  Single  & MTL-D  & MTL-I & Single  & MTL-D  & MTL-I\\\hline
        Element &  &&  &  &  &  &  &\\
        \quad Bearing & 81.00 &79.96& \textbf{81.27}&86.09&86.30& \textbf{88.59}&\textbf{93.20}&91.59& 90.77\\
        \quad Bracing &72.01 &77.32&\textbf{78.43} & 85.15&\textbf{88.65}&83.38 &82.35&85.82& \textbf{92.96}\\
        \quad Deck & 86.03&88.57&\textbf{91.05} & 95.31&\textbf{96.63}&96.55 &89.83&91.39 &\textbf{94.11}\\
        \quad Floor beam & 87.60 &88.09&\textbf{88.44} & 93.48&95.82&\textbf{97.09} &\textbf{93.30}& 91.60& 90.85\\
		\quad Girder & 94.07 &94.10&\textbf{94.65} &96.23&96.07&\textbf{96.68}&97.67&\textbf{97.87} &97.83\\
		\quad Substructure &82.41&84.04&\textbf{85.05} &87.31&88.05&\textbf{89.13}&93.62&84.86 &\textbf{94.89}\\
		\quad Background & 77.10 &78.87&\textbf{79.50} &91.67&92.53&\textbf{92.58} & 82.91&84.16 &\textbf{84.91}\\
		Defect&  & & & && &&\\
		\quad Corrosion &68.06 &\textbf{71.84}&70.49 &76.76&\textbf{82.68}&82.21&\textbf{85.75}&84.57 & 83.17\\
		\quad No corrosion & 93.64 &\textbf{94.77}&94.52 &\textbf{97.65}&97.50&97.27&95.80&\textbf{97.13}& 97.09\\\hline
		\end{tabular}
		\begin{tablenotes}
        \footnotesize
        \item \textit{Note}: The highest values of each metric for each class among all networks are shown in bold.
        \end{tablenotes}
        \end{threeparttable}
	\end{center}
\end{table*}

\subsection{Efficiency}

Besides the promising improvement of the MTL model on the task performance metrics, another advantage of MTL over the model composed of two independent single-task networks is efficiency. Table \ref{tab:time} compares the computational time in terms of training time and inference speed between the single-task-based model and a MTL model like MTL-D and MTL-I.

The total training time of the two single-task networks is about 6 minutes longer than that of the MTL-I network. However, it was almost double the MTL-D network training time. This is because the MTL-D network has quite similar architecture to the single-task networks except for an additional segmentation head. The MTL-L has a more complex architecture and projection matrix. The MTL models can reduce training time because, instead of investing time training many models on multiple tasks, they only need to train a single model with one or more model components shared by the multiple tasks. In this study, the two tasks share the feature extractor in a MTL model.

For the inference speed, the two MTL models have similar performance, which is faster than the single-task-based model. Conducting the forward pass through the encoder is the main computational overhead for the inference. Using the MTL model speeds up the inference due to sharing a deep feature extractor for the two tasks.
 
\begin{table}[!ht]
	\caption{Comparison of Computational Time}\label{tab:time}
	\begin{center}
	    \begin{threeparttable}
		\begin{tabular}{l r r r}
		\hline
			Computation time & Single & MTL-D & MTL-I  \\\hline
			Training time (min) & 26.85 & 15.93 &  20.42\\
			Inference speed\tnote{*} \text{ }(FPS) & 11.3 & 14.6 & 14.6  \\\hline 
		\end{tabular}
		\begin{tablenotes}
        \footnotesize
        \item[*]Nvidia Tesla V100 GPU with batch size 1.
        \end{tablenotes}
        \end{threeparttable}
	\end{center}
\end{table}

It is clear that MTL models outperform the single-task-based model on computational efficiency, not at the cost of sacrificing performance. Besides, the MTL model can provide a preliminary evaluation of bridge elements based on spatially associated results of element parsing and defect segmentation for the inspector. Specifically, outputs from the MTL model can be conveniently used to evaluate the proportion of corrosion area in each kind of element. The proportion can be a metric for the inspector to produce a preliminary condition assessment of bridge elements. It also can be a reference for selectively applying other high-resolution sensors to areas with developed surface corrosion. Using the two single-task networks needs extra post-processing effort to produce the desired condition assessment result. Therefore, the MTL model is more suitable than single-task networks for deploying on the UAV platform, which can save the limited computational resource on the device.

\subsection{Qualitative Examples}

\begin{figure*}[htbp]
    \centering
    \includegraphics[width=6in]{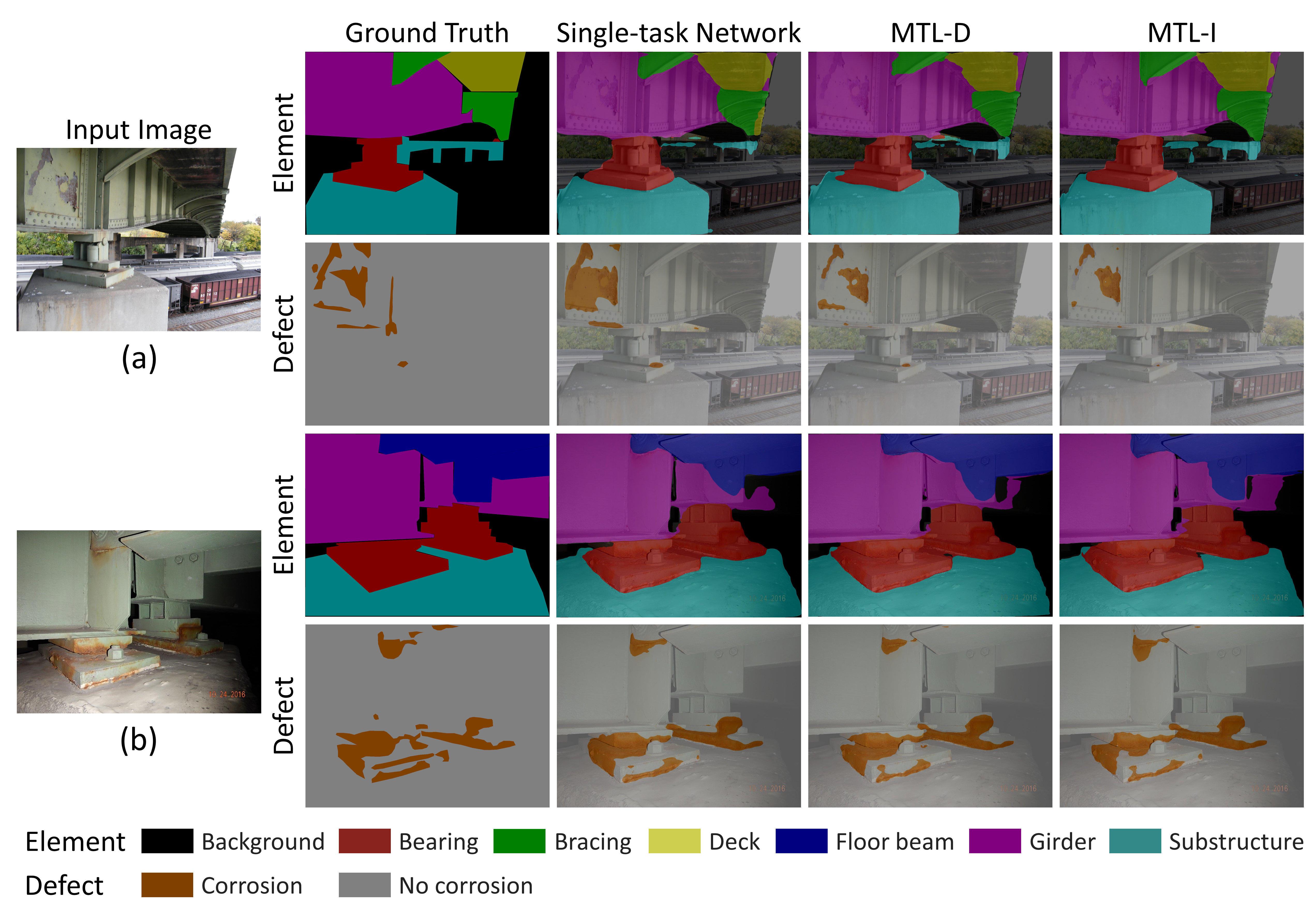}
    \caption{Comparison of the Results from Single-task Networks and MTL Networks}
    \label{fig:example}
\end{figure*}

Figure \ref{fig:example} illustrates two typical examples that implement two versions of the MTL model (D and L) and the single-task-based model, respectively. The MTL model performed better than single-task-based model on both tasks because the bridge element parsing task and corrosion segmentation task would cooperate to strengthen the prediction accuracy.

For the bridge parsing task, the MTL-I model had a better performance than others. In example (a), the single-task network for element parsing failed to segment the substructure in the distance and part of the bracing elements, and the MTL-I model segmented them successfully. Compared with the MTL-D model, MTL-I has fewer false positives for the deck and more true positives for the substructure. In example (b), the segmentation results from the single-task network and the MTL-D model on the floor beam were incomplete than MTL-I. The MTL-I model also generated more accurate segmentation results on the girder in the dark area of the input image. These observations explained why the MTL-I model obtained the highest precision and IoU on extracting the floor beam and girder among all the models compared in Table \ref{tab:class}.

For the corrosion segmentation task, the MTL-D model provided the most accurate segmentation result among all.
Specifically, the MTL-D model generated fewer false positives than the single-task network and more true positives than MTL-I on corrosion prediction, which suggests the reason for the MTL-D model achieving the highest precision on segmenting corrosion. In example (a), the MTL-D model is more accurate than the single-task and MTL-I models in segmenting the left bearing. In example (b), the MTL-D model provided a corrosion segmentation result similar to the single-task and MTL-L models. These examples indicate that the element parsing task and corrosion segmentation benefited from each other on edge between different classes.

\section{Conclusions}

This paper presented an MTL model for parsing bridge elements in bridge inspection images and simultaneously segmenting defects on the elements. Using a shared deep feature extractor, the proposed model learns features of both bridge elements and surface defects. Then, it split into two branches, each composed of a feature projection and a segmentation network. This study shows that training related tasks collaboratively helps boost the model's performance by better learning generalized and contextual information. The study also developed annotations of bridge elements and corrosion in the bridge inspection images to assess the performance of the MTL model. The effectiveness of the different designs for the model was explored, including the feature projection network, the cross-talk sharing between the network's two branches, and the loss function. The experimental evaluation on the dataset confirms that the proposed MTL model performs better than the single-task networks, with a 2.59\% improvement in the mIoU on the element parsing task and 2.46\% on the corrosion segmentation task. The MTL model also offers lower computational time with less training time and faster inference speed. The quantitative and qualitative results confirm that the MTL model has multiple advantages over single-task networks.

This paper has identified room for improvement. For example, further improving the computational efficiency to achieve real-time inference is highly desired when moving forward to real-world implementation. A recommendation system needs to be developed to make the proposed model an integrated tool for bridge inspection. Based on the preliminary result of the RGB image-based condition assessment, the recommendation system will suggest focused areas for high-resolution sensors to collect data. Model adaptation is also a critical future study. If new classes of bridge elements and defects need to be analyzed, few-shot class-incremental learning can efficiently bring those new classes to the model. For different types of bridges with no actual data captured by drones, synthetic training data of those bridges can be generated from their computer-aided design models and used for retraining the model. Moreover, the dataset needs to be extended with more images and annotations of bridge elements and defects to benefit the broader research community. As the research on this topic moves forward, the performance of MTL models will continue improving.


\begin{ac}
The authors confirm contribution to the paper as follows: study
conception and design: C. Zhang, M. M. Karim, R. Qin; data collection:
C. Zhang; analysis and interpretation of results: C. Zhang, R. Qin; draft manuscript preparation: C. Zhang, M. M. Karim, and R. Qin. All authors reviewed the results and approved the final version of the manuscript.
\end{ac}

\begin{dci}
The authors declared no potential conflicts of interest with respect to the research, authorship, and/or publication of this article. 
\end{dci}

\begin{funding}
The author(s) disclosed receipt of the following financial support for the research, authorship, and/or publication of this article: Qin and Karim are support by NSF through the award ECCS-2026357. Zhang and Karim are supported by NSF through the award ECCS-2025929. Any opinions, findings, and conclusions or recommendations expressed in this material are those of the authors and do not necessarily reflect the views of the National Science Foundation.
\end{funding}

\begin{das}
Data, models, and code that support the findings of this study are available at \url{https://github.com/itschenyu/Multitask-Learning-Bridge-Inspection}.
\end{das}

\bibliographystyle{TRR}
\bibliography{trb_template}

\begin{thebibliography}{10}
\providecommand{\url}[1]{\texttt{#1}}
\providecommand{\urlprefix}{URL }
\expandafter\ifx\csname urlstyle\endcsname\relax
  \providecommand{\doi}[1]{doi:\discretionary{}{}{}#1}\else
  \providecommand{\doi}{doi:\discretionary{}{}{}\begingroup
  \urlstyle{rm}\Url}\fi
\providecommand{\eprint}[2][]{\url{#2}}

\bibitem{ARTBA}
{American Road \& Transportation Builders Association ({ARTBA})}.
\newblock ARTBA Annual Bridge Report.
\newblock https://artbabridgereport.org, 2022.
\newblock Accessed August 1, 2022.

\bibitem{card}
{American Society of Civil Engineers (ASCE)}.
\newblock 2021 Report Card for America's Infrastructure.
\newblock https://infrastructurereportcard.org/cat-item/bridges-infrastructure,
  2021.
\newblock Accessed August 1, 2022.

\bibitem{inspection}
{Federal Highway Administration (FHWA)}.
\newblock National Bridge Inspection Standards, 2022.

\bibitem{spencer2019advances}
Spencer~Jr, B.~F., V.~Hoskere, and Y.~Narazaki.
\newblock Advances in Computer Vision-Based Civil Infrastructure Inspection and
  Monitoring.
\newblock \emph{Engineering}, Vol.~5, No.~2, 2019, pp. 199--222.

\bibitem{aashto2019manual}
{American Association of State Highway and Transportation Officials (AASHTO)}.
\newblock Manual for Bridge Element Inspection, 2nd Edition, 2019.

\bibitem{bianchi2022visual}
Bianchi, E. and M.~Hebdon.
\newblock Visual Structural Inspection Datasets.
\newblock \emph{Automation in Construction}, Vol. 139, 2022, p. 104299.

\bibitem{narazaki2020vision}
Narazaki, Y., V.~Hoskere, T.~A. Hoang, Y.~Fujino, A.~Sakurai, and B.~F.
  Spencer~Jr.
\newblock Vision-Based Automated Bridge Component Recognition with High-Level
  Scene Consistency.
\newblock \emph{Computer-Aided Civil and Infrastructure Engineering}, Vol.~35,
  No.~5, 2020, pp. 465--482.

\bibitem{bianchi2021coco}
Bianchi, E., A.~L. Abbott, P.~Tokekar, and M.~Hebdon.
\newblock COCO-Bridge: Structural Detail Data Set for Bridge Inspections.
\newblock \emph{Journal of Computing in Civil Engineering}, Vol.~35, No.~3,
  2021, p. 04021003.

\bibitem{karim2022semi}
Karim, M.~M., R.~Qin, G.~Chen, and Z.~Yin.
\newblock A Semi-Supervised Self-Training Method to Develop Assistive
  Intelligence for Segmenting Multiclass Bridge Elements from Inspection
  Videos.
\newblock \emph{Structural Health Monitoring}, Vol.~21, No.~3, 2022, pp.
  835--852.

\bibitem{zhang2022adeep}
Zhang, C., M.~M. Karim, and R.~Qin.
\newblock A Deep Neural Network for Multiclass Bridge Element Parsing in
  Inspection Image Analysis.
\newblock In \emph{Proceedings of the 8th World Conference on Structural
  Control and Monitoring}. 2022.

\bibitem{zhang2016road}
Zhang, L., F.~Yang, Y.~D. Zhang, and Y.~J. Zhu.
\newblock Road Crack Detection Using Deep Convolutional Neural Network.
\newblock In \emph{2016 IEEE International Conference on Image Processing)}.
  IEEE, 2016, pp. 3708--3712.

\bibitem{wang2020deep}
Wang, J., K.~Sun, T.~Cheng, B.~Jiang, C.~Deng, Y.~Zhao, D.~Liu, Y.~Mu, M.~Tan,
  X.~Wang, et~al.
\newblock Deep High-Resolution Representation Learning for Visual Recognition.
\newblock \emph{IEEE Transactions on Pattern Analysis and Machine
  Intelligence}, Vol.~43, No.~10, 2020, pp. 3349--3364.

\bibitem{yudin2020roof}
Yudin, D.~A., V.~Adeshkin, A.~V. Dolzhenko, A.~Polyakov, and A.~E. Naumov.
\newblock Roof Defect Segmentation on Aerial Images Using Neural Networks.
\newblock In \emph{International Conference on Neuroinformatics}. Springer,
  2020, pp. 175--183.

\bibitem{jia2021surface}
Jia, B., X.~Luo, R.~Tao, and Y.~Shi.
\newblock Surface Defect Detection of Aluminum Material Based on HRNet Feature
  Extraction.
\newblock In \emph{2021 4th International Conference on Data Science and
  Information Technology}. 2021, pp. 44--48.

\bibitem{akhyar2021beneficial}
Akhyar, F., C.-Y. Lin, and G.~S. Kathiresan.
\newblock A Beneficial Dual Transformation Approach for Deep Learning Networks
  Used in Steel Surface Defect Detection.
\newblock In \emph{Proceedings of the 2021 International Conference on
  Multimedia Retrieval}. 2021, pp. 619--622.

\bibitem{rubio2019multi}
Rubio, J.~J., T.~Kashiwa, T.~Laiteerapong, W.~Deng, K.~Nagai, S.~Escalera,
  K.~Nakayama, Y.~Matsuo, and H.~Prendinger.
\newblock Multi-Class Structural Damage Segmentation Using Fully Convolutional
  Networks.
\newblock \emph{Computers in Industry}, Vol. 112, 2019, p. 103121.

\bibitem{dung2019autonomous}
Dung, C.~V. et~al.
\newblock Autonomous Concrete Crack Detection Using Deep Fully Convolutional
  Neural Network.
\newblock \emph{Automation in Construction}, Vol.~99, 2019, pp. 52--58.

\bibitem{ronneberger2015u}
Ronneberger, O., P.~Fischer, and T.~Brox.
\newblock U-Net: Convolutional Networks for Biomedical Image Segmentation.
\newblock In \emph{International Conference on Medical Image Computing and
  Computer-assisted Intervention}. Springer, 2015, pp. 234--241.

\bibitem{pan2020automatic}
Pan, G., Y.~Zheng, S.~Guo, and Y.~Lv.
\newblock Automatic Sewer Pipe Defect Semantic Segmentation Based on Improved
  U-Net.
\newblock \emph{Automation in Construction}, Vol. 119, 2020, p. 103383.

\bibitem{wang2020unified}
Wang, M. and J.~C. Cheng.
\newblock A Unified Convolutional Neural Network Integrated with Conditional
  Random Field for Pipe Defect Segmentation.
\newblock \emph{Computer-Aided Civil and Infrastructure Engineering}, Vol.~35,
  No.~2, 2020, pp. 162--177.

\bibitem{zhao2017pyramid}
Zhao, H., J.~Shi, X.~Qi, X.~Wang, and J.~Jia.
\newblock Pyramid Scene Parsing Network.
\newblock In \emph{Proceedings of the IEEE Conference on Computer Vision and
  Pattern Recognition}. 2017, pp. 2881--2890.

\bibitem{shi2021improvement}
Shi, J., J.~Dang, M.~Cui, R.~Zuo, K.~Shimizu, A.~Tsunoda, and Y.~Suzuki.
\newblock Improvement of Damage Segmentation Based on Pixel-Level Data Balance
  Using VGG-Unet.
\newblock \emph{Applied Sciences}, Vol.~11, No.~2, 2021, p. 518.

\bibitem{he2017mask}
He, K., G.~Gkioxari, P.~Doll{\'a}r, and R.~Girshick.
\newblock Mask R-CNN.
\newblock In \emph{Proceedings of the IEEE International Conference on Computer
  Vision}. 2017, pp. 2961--2969.

\bibitem{ayele2020automatic}
Ayele, Y.~Z., M.~Aliyari, D.~Griffiths, and E.~L. Droguett.
\newblock Automatic Crack Segmentation for UAV-Assisted Bridge Inspection.
\newblock \emph{Energies}, Vol.~13, No.~23, 2020, p. 6250.

\bibitem{li2022multi}
Li, J., Q.~Wang, J.~Ma, and J.~Guo.
\newblock Multi-Defect Segmentation from Fa{\c{c}}ade Images Using Balanced
  Copy--Paste Method.
\newblock \emph{Computer-Aided Civil and Infrastructure Engineering}.

\bibitem{huang2022deep}
Huang, H., S.~Zhao, D.~Zhang, and J.~Chen.
\newblock Deep Learning-Based Instance Segmentation of Cracks from Shield
  Tunnel Lining Images.
\newblock \emph{Structure and Infrastructure Engineering}, Vol.~18, No.~2,
  2022, pp. 183--196.

\bibitem{yeum2019automated}
Yeum, C.~M., J.~Choi, and S.~J. Dyke.
\newblock Automated Region-of-Interest Localization and Classification for
  Vision-Based Visual Assessment of Civil Infrastructure.
\newblock \emph{Structural Health Monitoring}, Vol.~18, No.~3, 2019, pp.
  675--689.

\bibitem{liu2016ssd}
Liu, W., D.~Anguelov, D.~Erhan, C.~Szegedy, S.~Reed, C.-Y. Fu, and A.~C. Berg.
\newblock SSD: Single Shot Multibox Detector.
\newblock In \emph{European Conference on Computer Vision}. Springer, 2016, pp.
  21--37.

\bibitem{eigen2015predicting}
Eigen, D. and R.~Fergus.
\newblock Predicting Depth, Surface Normals and Semantic Labels with a Common
  Multi-Scale Convolutional Architecture.
\newblock In \emph{Proceedings of the IEEE International Conference on Computer
  Vision}. 2015, pp. 2650--2658.

\bibitem{bischke2019multi}
Bischke, B., P.~Helber, J.~Folz, D.~Borth, and A.~Dengel.
\newblock Multi-Task Learning for Segmentation of Building Footprints with Deep
  Neural Networks.
\newblock In \emph{2019 IEEE International Conference on Image Processing}.
  IEEE, 2019, pp. 1480--1484.

\bibitem{tian2015pedestrian}
Tian, Y., P.~Luo, X.~Wang, and X.~Tang.
\newblock Pedestrian Detection Aided by Deep Learning Semantic Tasks.
\newblock In \emph{Proceedings of the IEEE Conference on Computer Vision and
  Pattern Recognition}. 2015, pp. 5079--5087.

\bibitem{ruder2017overview}
Ruder, S.
\newblock An Overview of Multi-Task Learning in Deep Neural Networks.
\newblock \emph{arXiv preprint arXiv:1706.05098}.

\bibitem{zhang2021survey}
Zhang, Y. and Q.~Yang.
\newblock A Survey on Multi-Task Learning.
\newblock \emph{IEEE Transactions on Knowledge and Data Engineering}.

\bibitem{hoskere2020madnet}
Hoskere, V., Y.~Narazaki, T.~A. Hoang, and B.~Spencer~Jr.
\newblock MaDnet: Multi-Task Semantic Segmentation of Multiple Types of
  Structural Materials and Damage in Images of Civil Infrastructure.
\newblock \emph{Journal of Civil Structural Health Monitoring}, Vol.~10, No.~5,
  2020, pp. 757--773.

\bibitem{gong2019comparison}
Gong, T., T.~Lee, C.~Stephenson, V.~Renduchintala, S.~Padhy, A.~Ndirango,
  G.~Keskin, and O.~H. Elibol.
\newblock A Comparison of Loss Weighting Strategies for Multi Task Learning in
  Deep Neural Networks.
\newblock \emph{IEEE Access}, Vol.~7, 2019, pp. 141627--141632.

\bibitem{kendall2018multi}
Kendall, A., Y.~Gal, and R.~Cipolla.
\newblock Multi-Task Learning Using Uncertainty to Weigh Losses for Scene
  Geometry and Semantics.
\newblock In \emph{Proceedings of the IEEE Conference on Computer Vision and
  Pattern Recognition}. 2018, pp. 7482--7491.

\bibitem{Bianchi2021}
Bianchi, E. and M.~Hebdon.
\newblock Corrosion Condition State Semantic Segmentation Dataset, 2021.
\newblock \doi{10.7294/16624663.v2}.

\bibitem{russell2008labelme}
Russell, B.~C., A.~Torralba, K.~P. Murphy, and W.~T. Freeman.
\newblock LabelMe: a Database and Web-Based Tool for Image Annotation.
\newblock \emph{International Journal of Computer Vision}, Vol.~77, No.~1,
  2008, pp. 157--173.

\bibitem{pascal-voc-2012}
Everingham, M., L.~Van~Gool, C.~K.~I. Williams, J.~Winn, and A.~Zisserman.
\newblock The {PASCAL} {V}isual {O}bject {C}lasses {C}hallenge 2012 {(VOC2012)}
  {R}esults.
\newblock
  http://www.pascal-network.org/challenges/VOC/voc2012/workshop/index.html.

\bibitem{BharathICCV2011}
Hariharan, B., P.~Arbelaez, L.~Bourdev, S.~Maji, and J.~Malik.
\newblock Semantic Contours from Inverse Detectors.
\newblock In \emph{International Conference on Computer Vision (ICCV)}. 2011.

\end{thebibliography}

\end{document}